\documentclass[runningheads]{llncs}
\usepackage[T1]{fontenc}
\usepackage{graphicx,verbatim}
\usepackage{xcolor}
\usepackage{booktabs}
\usepackage{multirow}
\usepackage{amsmath}   
\usepackage{amssymb}   
\usepackage{esvect}
\usepackage[misc]{ifsym} 
\usepackage{cite}
\usepackage{hyperref}
\usepackage{color}
\usepackage{float}
\usepackage{caption}
\begin{document}
\title{Object Tokens as a Bridge Between Segmentation and Visual Question Answering in Robotic Surgery}
\titlerunning{Object Tokens as a Bridge Between Segmentation and VQA in surgery}
%

\author{Yiping Li\inst{1}$^{\textrm{\Letter}}$
\and Ronald de Jong\inst{1}
\and Romy van Jaarsveld\inst{3}
\and Franco Badaloni\inst{3}
\and Gino Kuiper\inst{3}
\and Jelle Ruurda\inst{3}
\and Josien Pluim\inst{1}
\and Marcel Breeuwer\inst{1,2}
}
\authorrunning{Y. Li et al.}
%
\institute{
Department of Biomedical Engineering, Eindhoven University of Technology, The Netherlands
\and
Department of Electrical Engineering, Eindhoven University of Technology, The Netherlands
\and
Department of Surgery, University Medical Center Utrecht, The Netherlands
\\
     \email{y.li9@tue.nl}}
  
\maketitle              
\begin{abstract}
Visual Question Answering (VQA) in robotic surgery, referred to as surgical VQA, requires high-level understanding of complex surgical scenes and the integration of visual perception with language reasoning, with the potential to support surgical training and intraoperative decision-making. Recent Vision–Language Models (VLMs) have shown promising performance through parameter-efficient fine-tuning; however, most existing approaches rely on coarse visual grounding, typically limited to bounding boxes, which fails to capture the fine-grained spatial structure of surgical objects. In this work, we propose a unified framework that jointly performs pixel-level segmentation and visual question answering within a single framework. Our approach integrates a VLM with a Segment Anything Model (SAM)-based decoder and represents scene elements as object tokens generated by the VLM. These object tokens guide answer prediction and are further projected to the SAM-based decoder to produce segmentation masks. By optimizing the object token embeddings through both segmentation and question answering objectives, the model learns spatially grounded representations that enhance visual reasoning while providing explicit pixel-level grounding. We evaluate the proposed method on the private RAMIE (Robot-Assisted Minimally Invasive Esophagectomy) dataset and the public EndoVis18 dataset, where it consistently outperforms baseline methods for surgical VQA. These results demonstrate that incorporating context-aware object tokens into vision–language models improves fine-grained surgical scene understanding. Code will be made publicly available upon acceptance.


\end{abstract}
\section{Introduction}
Trainees in surgical education often have limited access to expert mentorship and therefore rely on recorded procedures for observational learning \cite{seenivasan2022surgical}. However, passive video review does not allow for individualized, context-specific questioning. Surgical Visual Question Answering (surgical VQA) addresses this limitation by enabling natural language queries about robotic surgical scenes. The task requires high-level understanding of complex intraoperative visual data and the integration of visual perception with language reasoning, with the potential to enhance the accessibility and scalability of surgical training.

Building on the surgical VQA task, Visual Question Localized Answering for robotic surgery (Surgical-VQLA) extends question answering by localizing image regions that are most relevant to predicted question–answer pairs. Existing approaches typically employ Transformer-based backbones with cross-modal attention fusion, followed by regression-based bounding box heads and classification heads to jointly predict answers and spatial locations \cite{bai2023cat, bai2025surgical}. Subsequent work has explored architectural improvements, such as incorporating Mamba-based designs to enhance performance \cite{hao2025surgical}. In addition, \cite{hao2025enhancing} proposed leveraging large language model (LLM)–powered knowledge graphs to enhance localized question answering, representing an early step toward structured reasoning in this domain. While these methods demonstrate the ability to answer questions while providing spatial grounding, their grounding is limited to bounding boxes, and the language outputs are restricted to predefined classification categories rather than open-text generation.
 
More recently, surgical VQA has followed the broader trend of adopting foundational vision–language models (VLMs) that integrate visual encoders with LLMs to enable open-ended text generation. Both EndoChat \cite{WANG2026103789} and SurgVLM \cite{zeng2025surgvlm} introduce large-scale surgical datasets for this task. EndoChat adapts mixed pretrained visual encoders with an LLM using parameter-efficient fine-tuning via low-rank adaptation (LoRA) \cite{hu2022lora}, while SurgVLM further benchmarks multiple VLM architectures, identifying Qwen2.5-VL \cite{bai2025qwen2} as a strong baseline and proposing domain-specific adaptations. Despite these advances, most existing VLM-based approaches rely on tokenized bounding box coordinates for spatial grounding, which is insufficient for capturing fine-grained surgical structures, and explicit spatial reasoning in surgical question answering remains underexplored.

In parallel, language-guided segmentation has emerged as a promising research direction, leveraging the interpretability and reasoning capabilities of LLMs to guide pixel-level visual understanding. LISA \cite{lai2024lisa} first introduced this concept by extending the vocabulary with a special  \texttt{\textless seg\textgreater} token and proposing an embedding-as-mask paradigm to enable segmentation. This approach was subsequently extended by Vicas \cite{athar2025vicas}, which incorporated larger-scale video datasets with detailed human-written captions, temporally consistent annotations, and pixel-accurate masks for multiple objects with phrase grounding. CoVT \cite{qin2025chain} further refined training strategies by integrating explicit reasoning patterns to enhance model reasoning capabilities, while additional anchoring-based approaches have explored alternative embeddings, such as depth and edge cues, to strengthen vision-based reasoning.

\begin{figure}[htbp!]
\label{overview}
\centering
\includegraphics[width=0.95\textwidth]{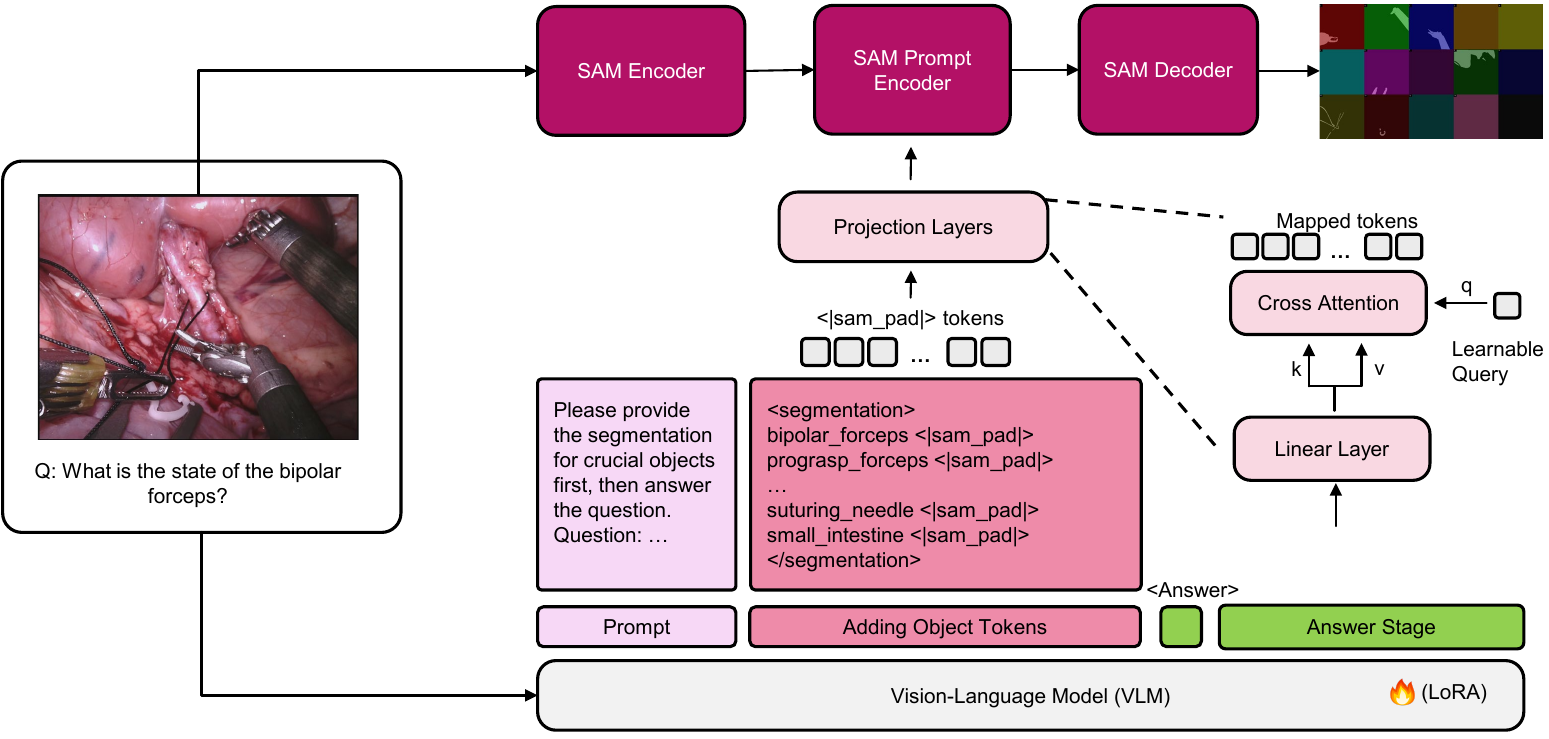}
\caption{Overview of the proposed method.}
\label{overview}
\end{figure}

Unlike general computer vision and language datasets, where reasoning can often be explicitly articulated in text, surgical question answering relies predominantly on visual reasoning. Answering a question requires understanding the surgical scene before generating a response. Motivated by this, we propose injecting object tokens representing crucial elements of the scene into VLM training. These tokens provide latent contextual primitives to enhance question answering and can be projected to a Segment Anything Model (SAM)-based decoder to produce segmentation masks. Our contributions are: (1) to the best of our knowledge, this is the first work to unify segmentation and question answering in the surgical domain, (2) we validate our approach on two surgical datasets and multiple SAM variants to demonstrate its effectiveness, and (3) we visualize attention maps to explore the interpretability of object tokens guided question answering.

\section{Methods}
Our method represents each segmentation target as a discrete object token in the VLM by appending a special \texttt{\textless sam\_pad\textgreater} token to the object class name. These tokens are treated as standard text tokens and trained with next-token cross-entropy loss. During training, the SAM decoder operates exclusively on the \texttt{\textless sam\_pad\textgreater} tokens to generate segmentation masks under ground-truth supervision, while the object tokens are jointly optimized through both question answering and segmentation objectives. At inference time, segmentation decoding is optional: the VLM can use the learned object tokens as contextual cues for answer generation, or project them into the SAM decoder to produce pixel-level masks when required. Segmentation performance depends on the quality of the VLM-generated object embeddings, the projection layers, and SAM components. An overview of the proposed framework is shown in Figure~\ref{overview}.

\subsubsection{Connecting SAM with VLM model}

We map each \texttt{\textless sam\_pad\textgreater} produced by the VLM into the prompt embedding space of SAM and use the SAM mask decoder for segmentation. While SAM \cite{kirillov2023segment} and SAM2 \cite{ravi2024sam} typically generate masks from prompt embeddings derived from points, bounding boxes, or masks, prior works \cite{athar2025vicas, lai2024lisa} have shown that the mask decoder can also operate directly on embeddings, bypassing explicit geometric prompts. We adopt this paradigm by using the projected object tokens as dense prompts; for SAM3 \cite{carion2025sam}, these projected tokens are instead treated as text prompts compatible with its architectural design. Regarding the projection layers, they are defined as follows.

We first project VLM latent features into the SAM decoder's prompt space using a single linear layer. This projection is formulated as
\begin{equation}
z_m = W z + b,
\end{equation}
where $z$ denotes the VLM latent feature \texttt{\textless sam\_pad\textgreater}, and $z_m$ is the mapped feature in the prompt space after the linear transformation.

Next, we introduce a set of learnable queries $q$, with the number of queries equal to the number of object classes to be segmented. The mapped feature $z_m$ serves as both the key $k$ and value $v$ in a cross-attention layer. The projected tokens are computed as
\begin{equation}
z_p = \mathrm{Attn}(q, k, v) = \mathrm{softmax}\left(\frac{q k^\top}{\sqrt{d_k}}\right) v,
\end{equation}
where $d_k$ is the dimension of the key vectors.

We then consider three versions of the SAM models for mask prediction, ensuring compatibility with each architecture. In the original SAM, each object mask is predicted by a mask decoder conditioned on a prompt token and the dense image embedding from the image encoder:
\begin{equation}
\hat{M}_i = \mathrm{Decoder}_{\mathrm{SAM}}(z_{p,i}, f), 
\quad \hat{M}_i \in \mathbb{R}^{H \times W},
\end{equation}
where $z_{p,i}$ is the projected prompt token corresponding to the $i$-th object, and $f$ denotes the dense image embedding. 
SAM2 extends this architecture by incorporating high-resolution, multi-scale features $\{f^{(l)}\}_{l=1}^{L}$ from the backbone to improve spatial precision:
\begin{equation}
\hat{M}_i = \mathrm{Decoder}_{\mathrm{SAM2}}\big(z_{p,i}, f, \{f^{(l)}\}_{l=1}^{L}\big), 
\quad \hat{M}_i \in \mathbb{R}^{H \times W}.
\end{equation}

Finally, SAM3 enables direct text-to-mask prediction by integrating a textual prompt embedding $t_i$ into a transformer encoder-decoder pipeline. The encoder fuses backbone features $f_{\mathrm{enc}}$ with the text embedding, and the decoder produces an intermediate representation:
\begin{equation}
h_i = \mathcal{T}_{\mathrm{dec}}\big(\mathcal{T}_{\mathrm{enc}}(f_{\mathrm{enc}}, t_i), t_i\big),
\end{equation}
which is subsequently processed by the segmentation head to produce the final mask:
\begin{equation}
\hat{M}_i = \mathcal{H}_{\mathrm{seg-SAM3}}(h_i, f_{\mathrm{enc}}, t_i), 
\quad \hat{M}_i \in \mathbb{R}^{H \times W}.
\end{equation}

Among all above, $\hat{M}_i$ denotes the predicted mask for the $i$-th object class. All SAM variants are trained with Dice loss and binary cross-entropy loss.

\subsubsection{Training Strategy and Curriculum Learning}

We adopt Qwen2.5-VL-7B \cite{bai2025qwen2} as the base VLM due to its strong performance on multimodal tasks. Parameter-efficient fine-tuning is performed using LoRA \cite{hu2022lora}, with a rank of 16 and a scaling factor ($\alpha$) of 32. We use a learning rate of $2\times10^{-4}$ and train for three epochs on each dataset.

Training is conducted using next-token cross-entropy loss, which supervises all tokens, together with the segmentation loss applied to the decoded masks, with the SAM components trained using a learning rate of $1\times10^{-5}$. In practice, multi-stage training is important to balance language modeling and segmentation objectives. We therefore employ a curriculum learning strategy: (i) initial warm-up using pure VQA data, (ii) training with segmentation supervision, and (iii) joint training with a mixture of pure VQA and segmentation-guided VQA.

\section{Experiments and Results}

\subsubsection{Datasets}
The EndoVis18 dataset was originally created for semantic segmentation of surgical images into tool and anatomical classes \cite{allan20202018} and has since become a widely used benchmark for surgical vision tasks. Here, we combine segmentation annotations from \cite{liu2025resurgsam2} with the expanded surgical VQA dataset from \cite{WANG2026103789}, which contains five question categories targeting different answer types. We follow the data split from \cite{WANG2026103789,bai2025surgical}, using videos 2–4, 6–7, 9–12, and 14–15 for training, and videos 1, 5, and 16 for testing.

\begin{figure}[bhtp!]
    \centering
    \includegraphics[width=\textwidth]{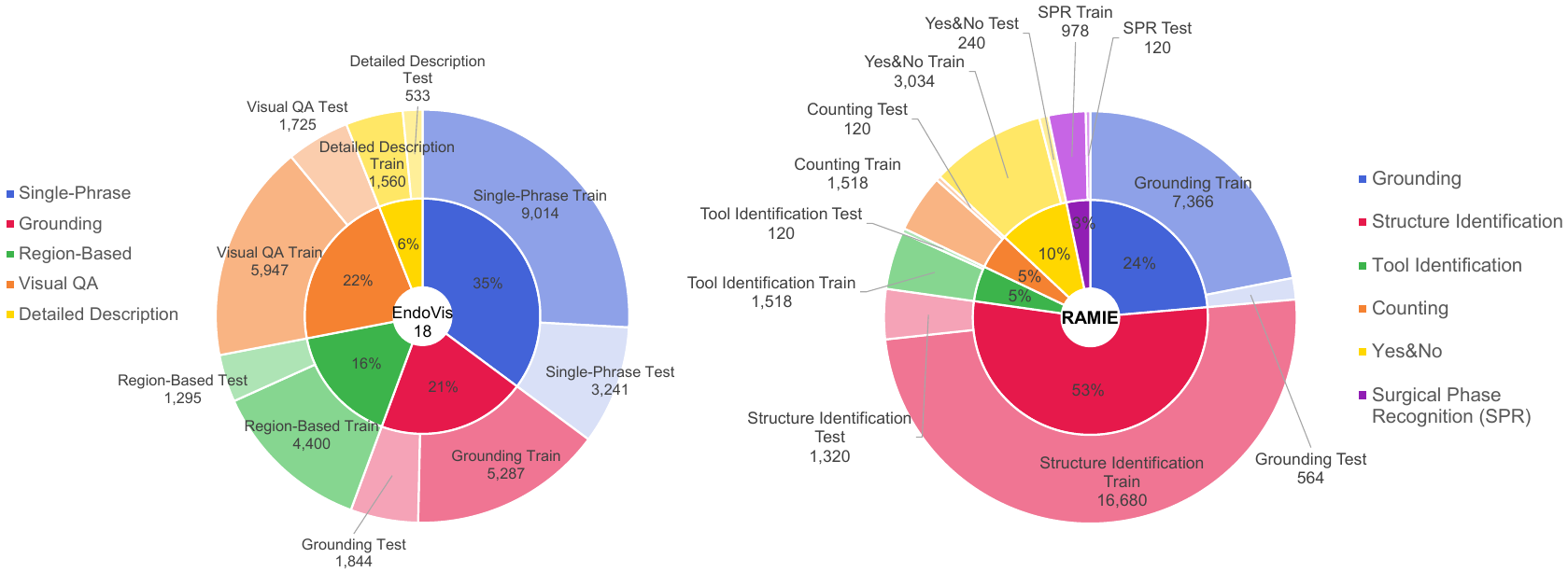}
    \caption{Dataset statistics for EndoVis18 (left) and RAMIE (right). The inner ring shows the percentage distribution per question category, while the outer ring displays the train–test split with absolute sample counts.}
    \label{fig:dataset}
\end{figure}
The Robot-Assisted Minimally Invasive Esophagectomy (RAMIE) dataset is an in-house collection of RAMIE procedure videos annotated with surgical phase labels and segmentation masks. A surgical VQA dataset is derived from these annotations. Similar to EndoVis18, the RAMIE surgical VQA dataset is also divided into multiple question categories. Statistics for both datasets are presented in Figure~\ref{fig:dataset}.

\subsubsection{Evaluation Metrics}
For surgical VQA task, we adopt the evaluation metrics used in EndoChat \cite{WANG2026103789}. Segmentation performance is evaluated using the Dice coefficient and the 95th percentile Hausdorff distance (HD95).

\subsubsection{Segmentation and VQA Results}
We evaluate segmentation performance by comparing our approaches with a strong segmentation baseline consisting of a DINOv3 ViT-L backbone pretrained on large-scale visual data \cite{simeoni2025dinov3}, coupled with an Encoder-only Mask Transformer (EoMT) \cite{kerssies2025your}, a recently proposed strong architecture for segmentation tasks. We assess our models' surgical VQA performance against the base VLM fine-tuned with LoRA, without additional \texttt{\textless sam\_pad\textgreater} tokens. Tables \ref{endovis_ablation} and \ref{ramie_ablation} present the comparative results on the two datasets.
\begin{table}[H]
\centering

\captionof{table}{Comparison experiment results on EndoVis18 dataset.}
\label{endovis_ablation}

\resizebox{\textwidth}{!}{%
\begin{tabular}{l cc c c ccc ccc c}
\hline
Model 
& \multicolumn{2}{c}{Segmentation}
& Single-Phrase 
& Grounding
& \multicolumn{3}{c}{Visual QA} 
& \multicolumn{3}{c}{Region-Based QA} 
& Description \\
\cline{2-3}\cline{4-4}\cline{5-5}\cline{6-8}\cline{9-11}\cline{12-12}
 & Dice & HD95
 & Acc 
 & mIoU 
 & BLEU-4 & CIDEr & Rouge-L 
 & BLEU-4 & CIDEr & Rouge-L 
 & GPT-4 \\
\hline

EoMT \cite{kerssies2025your}
& 0.50 & 54.19
& -
& -
& - & - & -
& - & - & -
& - \\

VLM only
& - & -
& 71.98
& 77.12
& 49.65 & 5.423 & 78.24 
& 58.87 & 5.042 & 79.04 
& 70.90 \\

VLM+SAM
& 0.44 & 39.66
& 75.12
& 74.72
& 52.55 & 5.986 & 79.43
& 60.27 & 5.588 & 83.11
& 68.80 \\

VLM+SAM2
& 0.52 & 30.65
& 73.85
& 78.35
& 53.16 & 5.811 & 80.88
& \textbf{62.94} & \textbf{6.651} & \textbf{83.69}
& 72.50 \\

VLM+SAM3
& \textbf{0.56} & \textbf{28.51}
& \textbf{76.67}
& \textbf{79.67}
& \textbf{54.80} & \textbf{6.049} & \textbf{81.57}
& 60.73 & 6.534 & 83.26
& \textbf{81.20} \\
\hline
\end{tabular}%
}

\vspace{1em}

\captionof{table}{Comparison experiment results on RAMIE dataset.}
\label{ramie_ablation}

\resizebox{\textwidth}{!}{%
\begin{tabular}{l cc c c c ccc ccc ccc}
\hline
Model 
& \multicolumn{2}{c}{Segmentation}
& Counting
& Yes\&No
& Grounding
& \multicolumn{3}{c}{Structure Identification} 
& \multicolumn{3}{c}{Tool Identification} 
& \multicolumn{3}{c}{Surgical Phase Recognition}  \\
\cline{2-3}\cline{4-4}\cline{5-5}\cline{6-6}\cline{7-9}\cline{10-12}\cline{13-15}
 & Dice & HD95
 & Acc
 & Acc 
 & mIoU 
 & BLEU-4 & CIDEr & Rouge-L 
 & BLEU-4 & CIDEr & Rouge-L 
 & BLEU-4 & CIDEr & Rouge-L  \\
\hline

EoMT \cite{kerssies2025your}
& 0.58 & 34.13
& - & -
& - & - & - & -
& - & - & -
& - & - & - \\

VLM only
& - & -
& 80.89 & 86.56 & 62.59
& 74.65 & 7.290 & 83.59 
& 90.43 & 8.192 & 95.03 
& 65.47 & 4.598 & 82.09 \\

VLM+SAM
& 0.51 & 36.62
& 80.58 & 85.58 & 61.58
& 74.14 & 7.053 & 82.25
& 90.25 & 7.995 & 94.58
& 59.34 & 4.126 & 81.08 \\

VLM+SAM2
& \textbf{0.61} & 34.14
& \textbf{83.33} & 87.92 & \textbf{64.78}
& \textbf{75.57} & \textbf{7.404} & \textbf{84.13}
& \textbf{91.34} & \textbf{8.389} & \textbf{96.02}
& \textbf{70.38} & \textbf{5.098} & \textbf{84.43} \\

VLM+SAM3
& 0.58 & \textbf{31.54}
& 81.67 & \textbf{89.16} & 64.44
& 75.32 & 7.351 & 83.81
& 90.82 & 8.355 & 95.79
& 68.62 & 4.978 & 83.70 \\
\hline
\end{tabular}%
}

\end{table}

These comparisons indicate that SAM-based segmentation achieves competitive performance relative to standard baselines. Among the evaluated variants, SAM2 generally outperforms the original SAM, likely due to its decoder design that leverages high-resolution image embeddings for finer mask prediction. Although SAM3 adopts a larger backbone and a more advanced architecture, it does not consistently outperform SAM2 in our experiments. This may be partly attributed to differences in optimization strategy. Specifically, while the official SAM3 training often employs multiple loss functions tailored to their training data characteristics, our experiments adopt a unified training protocol across all variants to ensure fair comparison, which may limit SAM3 performance under this simplified objective. Nevertheless, SAM3 still demonstrates strong segmentation capability, suggesting that increased model capacity can help capture fine-grained surgical visual details.

When integrating VLM with different SAM variants, we observe that VQA performance generally aligns with segmentation quality. Models with stronger segmentation backbones achieve higher VQA accuracy, indicating that the learned \texttt{\textless sam\_pad\textgreater} representations benefit the subsequent question-answering tasks. Incorporating object tokens with SAM2 and SAM3 further improves surgical VQA, particularly for surgical phase recognition in the RAMIE dataset, which relies on anatomical understanding, and for region-based QA in the EndoVis18 dataset, which focuses on specific anatomy or surgical tools.

\subsubsection{Comparison with SOTA models}
Our model demonstrates strong performance compared to fine-tuned VLM baselines as well as the in-domain surgical foundation model EndoChat \cite{WANG2026103789}, shown in Table \ref{endovis_sota}, highlighting its effectiveness in surgical VQA.

\begin{table}[htbp!]
\centering
\caption{Comparison with state-of-the-art models on EndoVis18 VQA dataset.}
\label{endovis_sota}
\resizebox{\textwidth}{!}{%
\begin{tabular}{l c c ccc ccc c}
\hline
Model 
& Single-Phrase 
& Grounded QA 
& \multicolumn{3}{c}{Visual QA} 
& \multicolumn{3}{c}{Region-Based QA} 
& Desc \\
\cline{2-2}\cline{3-3}\cline{4-6}\cline{7-9}\cline{10-10}
 & Acc 
 & mIoU 
 & BLEU-4 & CIDEr & Rouge-L 
 & BLEU-4 & CIDEr & Rouge-L 
 & GPT-4 \\
\hline

LLaMA-3.2-11B~\cite{grattafiori2024llama}
& 72.59 & 79.78 
& 49.87 & 4.549 & 78.36 
& 59.56 & 5.374 & 80.42 
& 63.52 \\

Gemma-3-12B~\cite{team2025gemma}
& 68.58 & 78.46 
& 52.15 & 5.919 & 78.36 
& 60.64 & 5.480 & 80.64 
& 65.32 \\

GLM-4V-9B~\cite{glm2024chatglm}
& 45.76 & 51.44 
& 37.42 & 3.160 & 54.47 
& 35.57 & 3.259 & 57.62 
& 57.86 \\

EndoChat~\cite{WANG2026103789}
& 71.47 & \textbf{86.89} 
& 52.20 & 5.990 & 79.62 
& 59.65 & 5.574 & 81.21 
& 79.35 \\

Ours
& \textbf{76.67}
& 79.67
& \textbf{54.80} & \textbf{6.049} & \textbf{81.57}
& \textbf{60.73} & \textbf{6.534} & \textbf{83.26}
& \textbf{81.20} \\
\hline

\end{tabular}%
}
\end{table}

\begin{figure}[ht!]
\centering
\includegraphics[height=6.5cm]{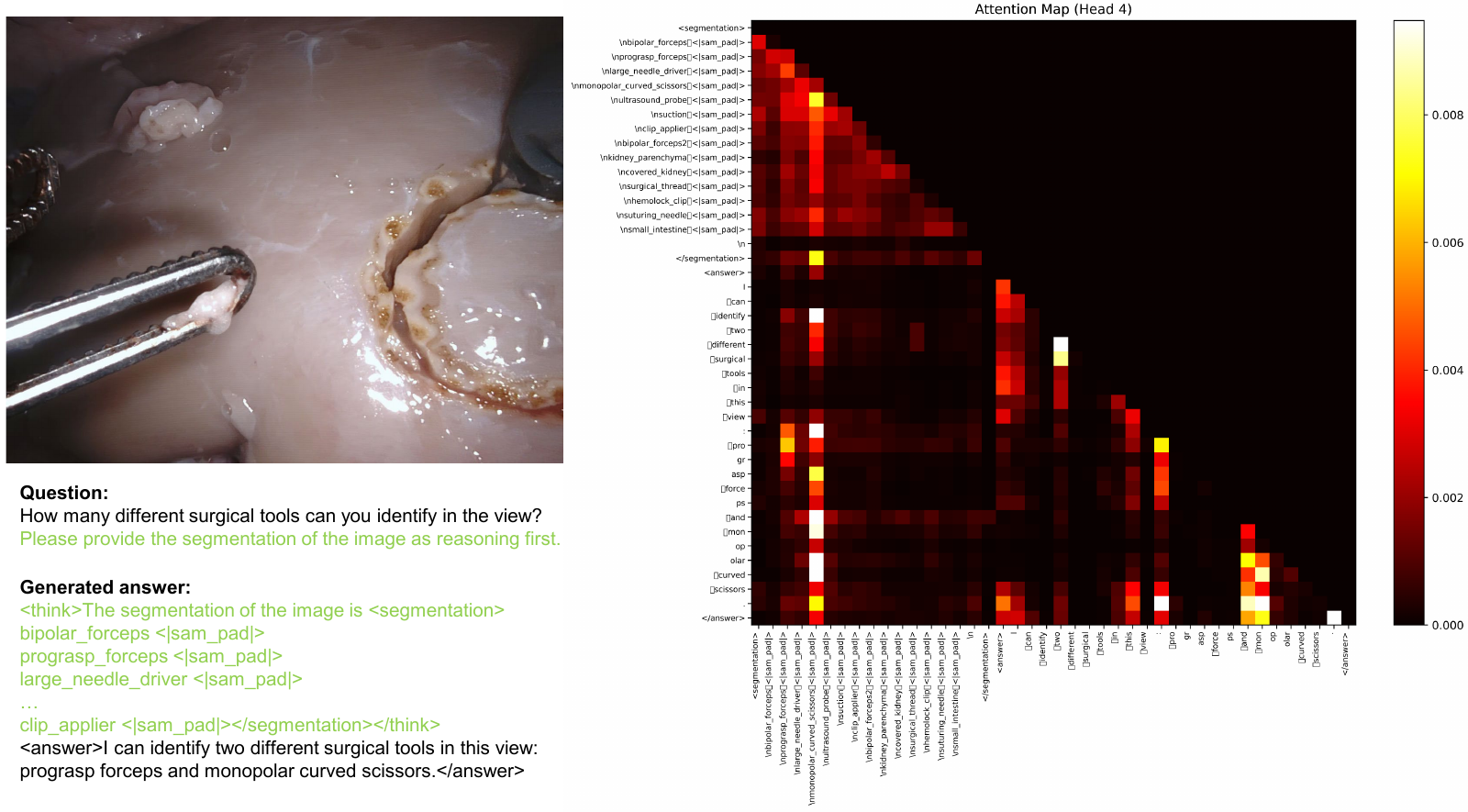}
\includegraphics[height=6.5cm]{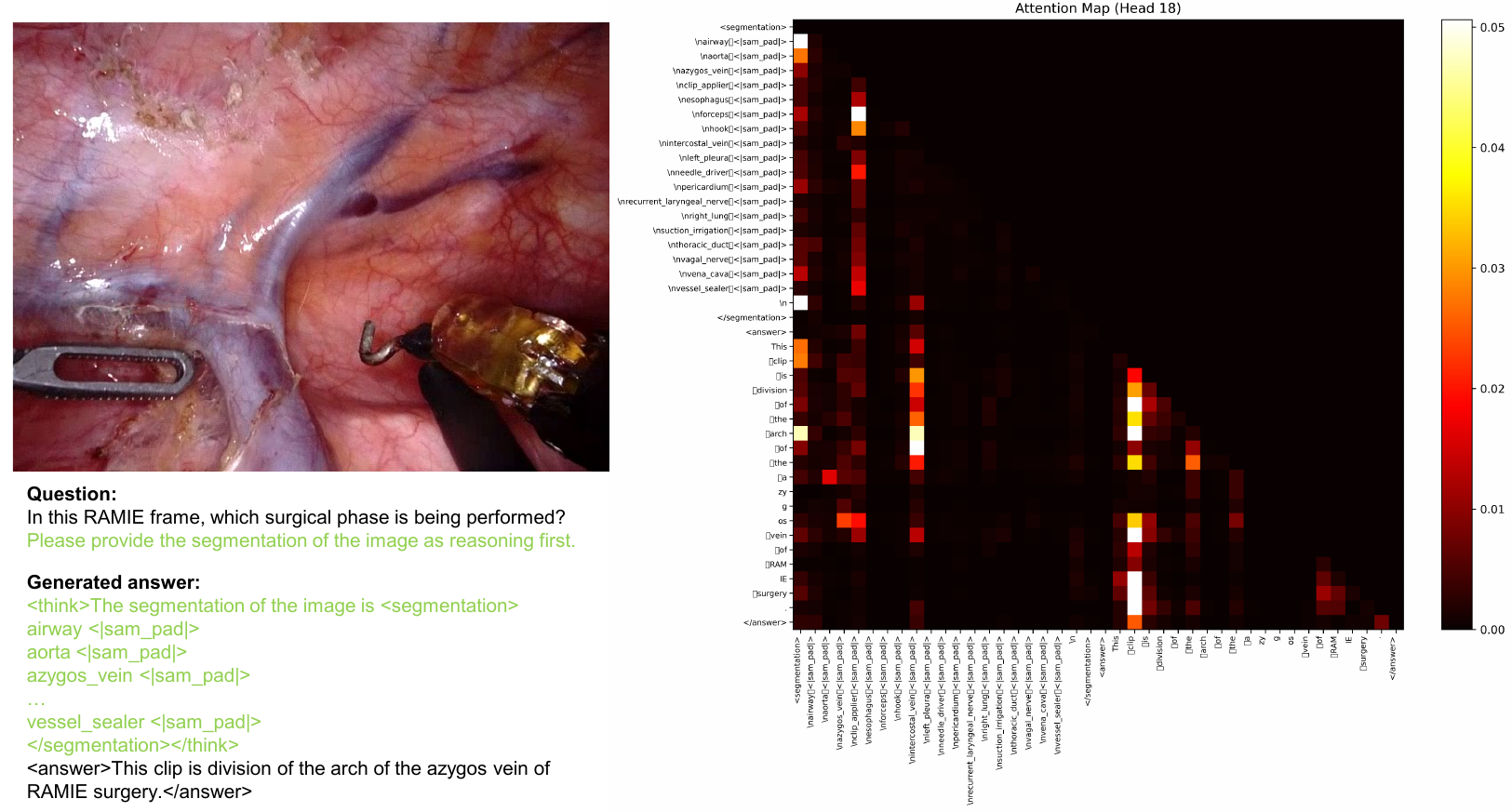}

\caption{Causal attention visualizations for object token generation and answer prediction. Top: example from the EndoVis18 dataset. Bottom: example from RAMIE dataset. Brighter regions indicate stronger attention weights.}
\label{attention_combined}
\end{figure}
\subsubsection{Attention Visualization}

We visualize the causal attention maps during object token generation and subsequent answer prediction. As this is a causal (autoregressive) attention map, each token can only attend to previously generated tokens. In Figure~\ref{attention_combined}, the horizontal axis represents the sequence of previously generated object tokens (read from left to right), while the vertical axis corresponds to the answer tokens generated at later time steps (read from top to bottom). Each cell therefore indicates how strongly a given answer token attends to an earlier object token. Brighter intensities denote higher attention weights.

The visualization shows that answer tokens assign higher attention to semantically relevant object tokens, indicating that the model conditions its predictions on grounded object representations. Although multiple transformer heads contribute to the final prediction—and some exhibit more uniformly distributed attention—we present representative heads in which attention is more concentrated, making the dependency structure more interpretable.
\section{Discussion and Conclusion}
We present a unified framework that jointly integrates pixel-level segmentation and open-text surgical VQA. By introducing learnable object tokens and jointly optimizing segmentation and question-answering objectives, the model learns meaningful object representations that enhance visual reasoning and can be decoded into explicit pixel-level segmentations. The framework is architecture-agnostic and can be adapted to different VLM backbones, offering flexibility for future extensions. While video data could provide richer context, publicly available surgical VQA datasets remain limited, so this work focuses on single-frame visual question answering. Existing datasets are also relatively shallow in reasoning complexity, emphasizing the need for clinically meaningful and diverse questions to fully evaluate model capabilities. Future work will extend this framework to video-based surgical VQA and investigate how combining segmentation with question answering can further improve spatially grounded reasoning in surgical scene understanding.

%
%
%
\bibliographystyle{splncs04}
\bibliography{ref}
%




\end{document}